\documentclass{ecai} 

\usepackage{latexsym}
\usepackage{amsthm}
\usepackage{booktabs}
\usepackage{enumitem}
\usepackage{graphicx}
\usepackage{xcolor}
\usepackage[dvipsnames]{xcolor}
\usepackage{amssymb, amsmath}
\usepackage{minted}
\usepackage{tikz}
\usetikzlibrary{calc}
\usepackage{multirow}
\usetikzlibrary{shapes}
\usetikzlibrary{positioning}
\usepackage{fixmath}
\usepackage{mathtools}
\usepackage{soul}
\usepackage{lipsum}
\usepackage{listings}
\usepackage{lipsum}
\usepackage{dsfont}

\newcommand{\deepproblog}{\textsc{DeepProbLog}}
\newcommand{\deepgraphlog}{\textsc{DeepGraphLog}}
\newcommand{\problog}{\textsc{ProbLog}}

\newcommand{\cF}{\mathcal{F}}
\newcommand{\cG}{\mathcal{G}}

\newcommand{\cR}{\mathcal{R}}

\newcommand{\BibTeX}{B\kern-.05em{\sc i\kern-.025em b}\kern-.08em\TeX}

\begin{document}


\begin{frontmatter}

\paperid{123} 


\title{\deepgraphlog{} for Layered Neurosymbolic AI}


\author[A,B]{\fnms{Adem}~\snm{Kikaj}\thanks{Corresponding Author. Email: adem.kikaj@kuleuven.be.}}
\author[A,B]{\fnms{Giuseppe}~\snm{Marra}}
\author[D]{\fnms{Floris}~\snm{Geerts}} 
\author[A,B]{\fnms{Robin}~\snm{Manhaeve}} 
\author[A,B,C]{\fnms{Luc}~\snm{De Raedt}} 

\address[A]{KU Leuven, Belgium}
\address[B]{Leuven.AI, Belgium}
\address[C]{Örebro University, Sweden}
\address[D]{University of Antwerp, Belgium}


\begin{abstract}
Neurosymbolic AI (NeSy) aims to integrate the statistical strengths of neural networks with the interpretability and structure of symbolic reasoning. However, current NeSy frameworks like \deepproblog{} enforce a fixed flow where symbolic reasoning always follows neural processing. This restricts their ability to model complex dependencies, especially in irregular data structures such as graphs. In this work, we introduce \deepgraphlog{}, a novel NeSy framework that extends \problog{} with \textit{Graph Neural Predicates}. \deepgraphlog{} enables multi-layer neural-symbolic reasoning, allowing neural and symbolic components to be layered in arbitrary order. In contrast to \deepproblog{}, which cannot handle symbolic reasoning via neural methods, \deepgraphlog{} treats symbolic representations as graphs, enabling them to be processed by Graph Neural Networks (GNNs). We showcase the capabilities of \deepgraphlog{} on tasks in planning, knowledge graph completion with distant supervision, and GNN expressivity. Our results demonstrate that \deepgraphlog{} effectively captures complex relational dependencies, overcoming key limitations of existing NeSy systems. By broadening the applicability of neurosymbolic AI to graph-structured domains, \deepgraphlog{} offers a more expressive and flexible framework for neural-symbolic integration. 

\end{abstract}

\end{frontmatter}


\section{Introduction}
The field of neurosymbolic AI (NeSy) aims to integrate learning and reasoning into a unified framework. Most existing NeSy systems process \texttt{subsymbolic} data with a \texttt{neural} component and pass the resulting \texttt{symbolic} output to a reasoning module—resulting in a fixed \texttt{subsymbolic $\xrightarrow{}$ neural $\xrightarrow{}$ symbolic} structure \cite{garcez2023neurosymbolic,hitzler2022neuro,marra2024statistical,van2021modular}.

A major limitation of this structure is the restricted interaction between neural and symbolic components during inference. Neural networks are typically applied to perception tasks and cannot operate directly on incomplete symbolic data. At the same time, symbolic modules cannot exploit the internal representations learned by neural models, limiting their ability to support integrated reasoning.

However, many tasks demand iterative reasoning, where symbolic representations are repeatedly refined by neural components. Reasoning may start from an incomplete \texttt{symbolic} input, refined by a \texttt{neural} module into a new \texttt{symbolic} form—yielding a \texttt{symbolic $\xrightarrow{}$ neural $\xrightarrow{}$ symbolic} flow \cite{van2021modular}. Enabling learning in such settings requires that both neural and symbolic components remain differentiable, allowing gradients to propagate across multiple stages.

\begin{figure*}[t]
    \begin{center}
    \begin{minipage}{0.45\textwidth}
        \centering
        \includegraphics[width=0.9\linewidth]{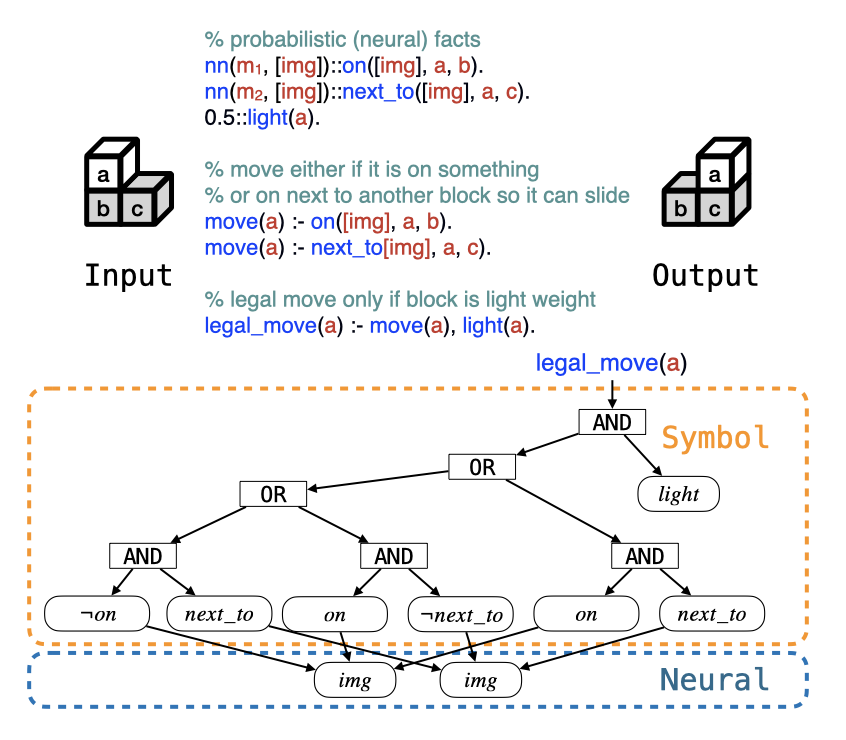}   
        
        {\textbf{(a)} \texttt{subsymbolic $\xrightarrow{}$ neural $\xrightarrow{}$ symbolic}}
        \label{fig:circuitA}
    \end{minipage}
    \hfill
    \begin{minipage}{0.45\textwidth}
        \centering
        \includegraphics[width=0.9\linewidth]{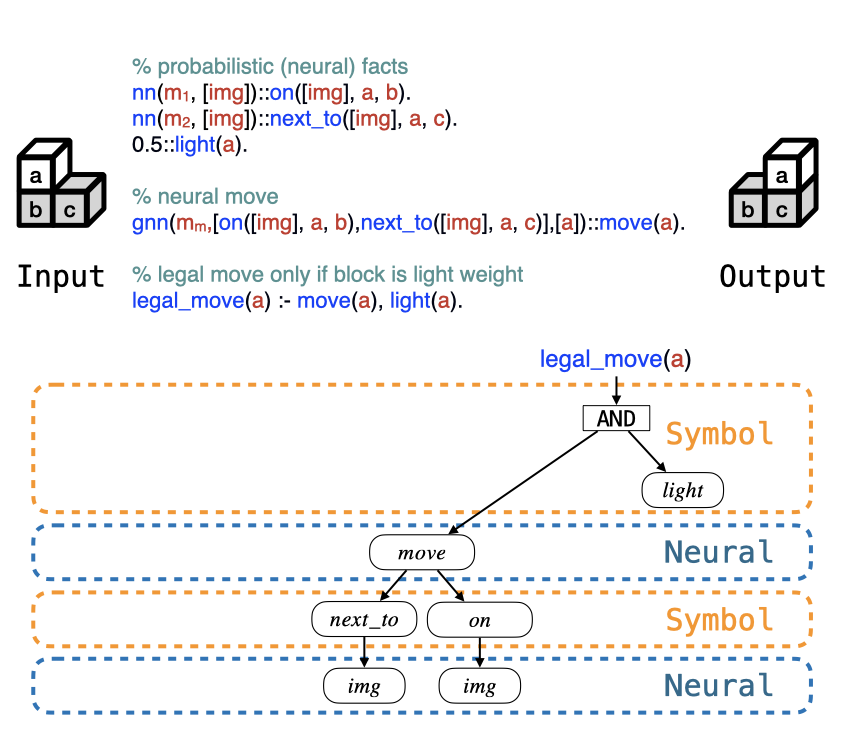}   
        
        {\textbf{(b)} \texttt{symbolic $\xrightarrow{}$ neural $\xrightarrow{}$ symbolic}}
        \label{fig:circuitB}
    \end{minipage}
    \end{center}
    \caption{Comparison of two reasoning paradigms: (a) standard \texttt{subsymbolic $\rightarrow$ neural $\rightarrow$ symbolic} flow, and (b) layered \texttt{symbolic $\rightarrow$ neural $\rightarrow$ symbolic} reasoning, as used in \deepgraphlog{}. }

    \label{fig:CircuitDiagram}
\end{figure*}

Figure \ref{fig:CircuitDiagram} presents two reasoning paradigms in a Blocks World-inspired task, where, for simplicity, the objective of the task is to move a single block. In part (a), the system follows a \texttt{subsymbolic $\rightarrow$ neural $\rightarrow$ symbolic} setting: visual input is processed by neural networks to extract symbolic relations (such as \texttt{on} and \texttt{next\_to}), which are then passed to a symbolic module that determines whether a move like \texttt{legal\_move(a)} is valid. In part (b), the system still begins with subsymbolic input, but reasoning unfolds in multiple alternating stages. Neural components operate not just at the perceptual level but also on intermediate symbolic structures (e.g., \texttt{on}, \texttt{next\_to}) to infer higher-level relations (e.g., \texttt{move(a)}), which are then used by symbolic modules for further reasoning. The elements depicted in the figure—such as neural facts and symbolic relations—will be introduced and explained in detail later in the paper.

The configuration in Figure \ref{fig:CircuitDiagram}(b) highlights a limitation of conventional NeSy approaches: their lack of support for recursive, layered reasoning where neural and symbolic modules interact across multiple inference steps. Allowing symbolic representations to be dynamically refined by neural inference makes it possible to reason effectively even in the presence of incomplete knowledge.

Overcoming this limitation requires neural components that can reason over symbolic structures—often incomplete, relational, and irregular—rather than only processing raw subsymbolic inputs. This motivates the use of Graph Neural Networks (GNNs), which have emerged as powerful architectures for structured learning and neural-based reasoning \cite{barcelo2020logical,nunn2024logic}. Their ability to model graph-structured data makes them well-suited for layered NeSy systems, where learning and reasoning operate over dynamic, symbolic representations. However, GNNs have not yet been deeply integrated into NeSy systems to support multi-layer, bidirectional interaction between neural and symbolic reasoning stages \cite{zhang2020efficient,vsourek2021beyond}.

To address the lack of a multi-layer interaction between neural and symbolic components in existing NeSy systems, we introduce \deepgraphlog{}, a new framework that integrates graph neural networks with symbolic reasoning in a fully declarative language. Building upon \deepproblog{} \citep{manhaeve2018deepproblog}, which connects neural networks to probabilistic logic programming via \textit{neural predicates}, our work extends this idea to \textit{graph neural predicates}. These predicates not only process subsymbolic data but also accept graph structures, enabling reasoning to happen at any layer in the model.

Our main contributions are (i) \textbf{A new neurosymbolic language}, \deepgraphlog{}, that integrates graph neural predicates within the probabilistic logic programming language \problog{} enabling neural and symbolic components to interact in a bidirectional way across multiple levels of a NeSy model; (ii) \textbf{A new learning setting for GNNs} that supports weakly supervised learning and integrating constraints; (iii) \textbf{A set of experiments} that demonstrate the flexibility and advantages of \deepgraphlog{} on tasks that combine graph-based learning with structured reasoning, highlighting its potential to solve problems that current NeSy systems cannot.


\section{Preliminaries}\label{sec:preliminaries}

\paragraph{Graphs}
A directed multi-relational labelled graph $G$ is a triple $(V(G), E(G), \ell_V)$, where $V(G)$ is a set of vertices, $E(G) \subseteq V(G) \times \Sigma_E \times V(G)$ is a set of labelled edges with $\Sigma_E$ a set of edge labels, and $\ell_V \colon V(G) \to 2^{\Sigma_V}$ assigns a set of vertex labels from a set $\Sigma_V$ to each vertex. Each edge $(u, \ell, v) \in E(G)$ connects two vertices $u$ and $v$ with a label $\ell \in \Sigma_E$, allowing multiple labelled edges between the same pair of vertices. Each vertex $v \in V(G)$ may have multiple labels, encoded as a subset $\ell_V(v) \subseteq \Sigma_V$.

\paragraph{Logic Programming}
We summarise basic concepts of logic programming. An in-depth discussion can be found in \cite{riguzzi2023foundations,baral1994logic}.

A \emph{term} $t$ is either a constant $c$, a variable $V$, or a structured term $f(t_1, \ldots, t_n)$, where $f$ is a function symbol and the $t_i$ are terms.
An \emph{atom} $a$ is an expression of the form $q(t_1, \ldots, t_n)$, where $q$ is a predicate symbol (of arity $n$, or $q/n$ for a shorthand notation) and $t_i$ are terms.
A \emph{literal} is either an atom or the negation $\lnot a$ of an atom. 

An \emph{expression} $e$ is a term, an atom, or a literal. An expression is \emph{ground} if it does not contain any variables. A \emph{substitution} $\theta = \{V_1 \mapsto t_1, \ldots, V_n \mapsto t_n\}$ assigns terms $t_i$ to variables $V_i$. A substitution is \emph{ground} if all terms $t_i$ are ground, i.e., contain no variables. Applying a substitution $\theta$ to an expression $e$, denoted $e\theta$, means replacing each occurrence of $V_i$ in $e$ with $t_i$. Ground substitutions turn an expression into a ground expression. Two expressions $e_1$ and $e_2$ \emph{unify} if there exists a substitution $\theta$ such that $e_1\theta = e_2\theta$. We assume that the domain of variables is finite, so that the set of possible ground substitutions is finite.

A \emph{rule} is a statement of the form $h \mathbin{:\! -} b_1, \ldots, b_n$, where $h$ is an atom (the \emph{head}) and each $b_i$ are literals (the \emph{body}).
The notation $\mathbin{:\! -}$ represents logical implication ($ \xleftarrow{}$), and the comma (\texttt{,}) represents conjunction ($\wedge$).
A rule with an empty body ($n = 0$) is called a \emph{fact}. A \emph{rule} is \emph{ground} if all terms in its head and body are ground. A \emph{ground instance} of a rule is obtained by applying a ground substitution—i.e., a substitution that replaces all variables with ground terms—resulting in a ground rule. Similarly, a \emph{ground fact} is a fact whose atom is ground.
A \emph{logic program} is a set of rules $\cR$. 

With $\Omega$ we denote the set of all ground atoms formed from the predicate symbols and ground terms. We assume that the set of function and constant symbols is finite, so that $\Omega$ is also finite. A \emph{possible world} $\omega$ is a subset $\omega \subseteq \Omega$.
A possible world $\omega$ \emph{satisfies} a ground instance of a rule $h \mathbin{:\!-} b_1, \ldots, b_n$ if, whenever all body literals $b_i$ are true in $\omega$, the head $h$ is also true in $\omega$. We say that $\omega$ \emph{satisfies} a logic program $\mathcal{R}$ if it satisfies every ground instance of every rule in $\mathcal{R}$; in that case, we write $\omega \models \mathcal{R}$ and say that $\omega$ is a \emph{model} of $\mathcal{R}$.

If every model $\omega$ of a logic program $\mathcal{R}$ also satisfies a ground atom $a$, we write $\mathcal{R} \models a$ and say that $a$ is a \emph{logical consequence} of $\mathcal{R}$, or that $\mathcal{R}$ \emph{entails} $a$.   
Among all models, the \emph{minimal model} is the model that is minimal with respect to set inclusion. The minimal model of a logic program $\mathcal{R}$, which we denote by $\omega_{\mathcal{R}}$, can be defined as $\omega_{\mathcal{R}} = \{ a\theta \mid \mathcal{R} \models a\theta \text{ and } a\theta \text{ is ground} \}$, it is, the set of all ground atoms that are logical consequences of $\mathcal{R}$. This possible world corresponds to the unique least model of $\mathcal{R}$.

\paragraph{Graphs in Logic Programming}
Notice that if we restrict our attention to unary and binary predicates and assume a function-free setting, a possible world $\omega$ can be naturally interpreted as a graph. The set of vertices corresponds to the set of constants, and edges (or labelled edges) correspond to the ground atoms involving those constants.
Each binary ground atom $q(c_1, c_2) \in \omega$ can be interpreted as a labelled edge $(c_1,q,c_2)$ from vertex $c_1$ to vertex $c_2$, where the predicate symbol $q \in \Sigma_E$ defines the edge label. Similarly, each unary ground atom $q(c) \in \omega$ can be interpreted as a label assigned to vertex $c$, where $q \in \Sigma_V$ defines the vertex label. Thus, a possible world encodes a directed multi-relational labelled graph, with vertex and edge labels derived from the predicate symbols.
The direction of edges inherited from the logical structure plays an important role during message passing in GNNs: messages are propagated along the directed edges, from the source node to the target node. If edge direction is not intended to influence the reasoning (i.e., if edges are to be treated as undirected), then both $q(c_1, c_2)$ and $q(c_2, c_1)$ would need to be present in $\omega$, ensuring bidirectional message flow between $c_1$ and $c_2$.

\section{The \deepgraphlog{} Framework}

We now introduce \deepgraphlog{}, a neurosymbolic framework that tightly integrates Graph Neural Networks (GNNs) with probabilistic logic programming, enabling layered reasoning over symbolic and subsymbolic data. As our approach is compatible with any GNN architecture, we do not fix or detail a specific architecture. \deepgraphlog{} extends \problog{}—by itself a \emph{probabilistic} extension of logic programs—by introducing \textit{graph neural predicates}.
We first recall the basics of \problog{} and illustrate them using the Blocks World examples shown in Figure \ref{fig:CircuitDiagram}. We then motivate the need for \deepgraphlog{} by highlighting the limitations of both \problog{} and \deepproblog{}. Although \deepproblog{} extends \problog{} with neural predicates for processing subsymbolic data, its semantics ultimately reduces to that  of \problog{}. For this reason, we directly extend \problog{} with graph neural predicates. At the same time, \deepgraphlog{} retains full support for the functionality of \deepproblog{}, including the use of neural predicates to handle subsymbolic inputs. Next, we introduce the syntax and semantics of \deepgraphlog{}, followed by a discussion on inference.

\subsection{\problog}\label{sec:ProbLog}
A \problog{} program \cite{de2007problog} extends a logic program $\cR$ by allowing some ground facts to be annotated with probabilities. A probabilistic fact is of the form $\texttt{p::f}$, where $p \in [0,1]$ is the probability that the ground atom $f$ holds. Formally, a \problog{} program is a pair $(\mathcal{F}, \mathcal{R})$, where $\cF$ is a set of ground probabilistic facts and $\cR$ is a set of rules. For example, the following \problog{} program is based on the task shown in Figure \ref{fig:CircuitDiagram} where we replace the neural predicates with probabilistic values:

\begingroup
\footnotesize
\begin{minted}[frame=single]{prolog}
0.7::on(a,b). 0.4::next_to(a,c). 0.5::light(a).
move(X) :- on(X,Y). move(X) :- next_to(X,Y).
legal_move(X) :- move(X), light(X).
\end{minted}
\endgroup
In this program, the facts \texttt{on(a,b)} and \texttt{next\_to(a,c)} are assigned probabilities, representing uncertainty about the position of block \texttt{a}. Additionally, the unary predicate \texttt{light(a)} expresses whether block \texttt{a} is light enough to move. The rules state that a block \texttt{X} is movable if it is either on another block or adjacent to one. A \texttt{legal\_move} is allowed only if the block is both movable and light.

In \problog{}, a possible world $\omega$ is defined as the minimal model of a set of sampled ground probabilistic facts together with rules $\cR$ following the semantics introduced in Section \ref{sec:preliminaries}. Formally, every subset $F \subseteq \cF$ defines a possible world $\omega_F = F \cup \{f\theta \mid \cR \cup F \models f\theta \text{ and } f\theta \text{ is ground}\}.$ 
For example, if $F = \{\texttt{on(a,b),light(a)}\}$, and the rules $\cR$ entail that \texttt{move(a)} and \texttt{legal\_move(a)} hold, then the resulting possible world $\omega_F$ is $\omega_F = \{\texttt{on(a,b),light(a)}\} \cup \{\texttt{move(a),legal\_move(a)}\}$ where both \texttt{move(a)} and \texttt{legal\_move(a)} are derived by applying the rules with the substitution $\theta = {X \mapsto \texttt{a}}$.
The probability of a possible world $\omega_F$ is computed by assuming independence of the ground probabilistic facts. Specifically, the probability is:
\begin{equation}\label{eq:ProbPossibleWorld}
    P(\omega_F) = \prod_{f \in F} P(f) \prod_{f \in \mathcal{F} \setminus F} (1 - P(f)),
\end{equation}

where $P(\texttt{f})$ is the probability assigned to the ground probabilistic fact $\texttt{f}$. For instance, the probability of the world where \texttt{on(a,b)} and \texttt{light(a)} are sampled, but \texttt{next\_to(a,c)} is not, is $ P(\omega_{\{\texttt{on(a,b),light(a)}\}}) = 0.7 \times 0.5 \times (1-0.4) = 0.21.$ Given that a \problog{} program defines a probability distribution over possible worlds, two fundamental inference tasks are supported:
First, the \emph{marginal probability} of a query computes the probability that a ground atom $q$ holds.  Formally, it is the sum of the probabilities of all possible worlds $\omega$ such that $\omega \models q$: $P(q)=\sum_{\omega_F\models q,\,F\subseteq\mathcal{F}}P(\omega_F)$.

Second, the \emph{conditional probability} of a query given evidence computes the probability that a ground atom $q$ holds, conditioned on another ground atom $e$ (the evidence) being true. The conditional probability is formally defined as: $P(q\,|\,e)=\dfrac{P(q \land e)}{P(e)}$.

\textbf{Random Graphs:} A random graph is a graph whose structure is determined by a probability distribution — typically, each edge exists independently with some probability. In this sense, a \problog{} program that includes probabilistic binary facts—such as \texttt{0.7::link(a,b)}—defines a distribution over possible graphs: each possible world corresponds to a graph in which some edges are present and others are not. This perspective aligns with the notion of random graphs and allows one to reason about their properties using logical rules. For instance, given a set of probabilistic \texttt{link/2} facts, one can define a recursive \texttt{path/2} predicate and query the probability that a path exists between two nodes (see \cite{de2007problog}).

\subsection{\deepgraphlog{}} 
In probabilistic reasoning, conditional dependencies describe how one event influences another. In \problog{}, these dependencies are introduced explicitly through rules, while probabilistic facts remain marginally independent. This resembles Bayesian networks, where root nodes are independent and edges encode dependencies. However, encoding such structure in \problog{} requires manually specifying all relevant rules and knowing the full dependency structure in advance, which quickly becomes impractical in complex domains \cite{de2015inducing}. In multi-layer reasoning, \problog{} operates in a \texttt{symbol $\rightarrow$ symbol $\rightarrow$ symbol} setting, where symbolic layers are stacked. While this supports logical composition, it is only feasible when the entire reasoning process is explicitly defined in advance. For example, Figure \ref{fig:CircuitDiagram} illustrates a task where rules define dependencies: a block can be moved if it is stacked on or next to another block. As dependencies grow, the number of configurations—and thus required ground rules—increases exponentially. Modelling such cases in \problog{} is only feasible when both the dependency structure and associated probabilities are fully known.

\deepproblog{} \cite{manhaeve2018deepproblog} extends \problog{} by introducing \emph{neural predicates}, allowing probabilistic facts to be parametrized by neural networks. However, it inherits the assumption that facts are marginally independent—preventing neural models from learning dependencies between them. Additionally, while \deepproblog{} can handle symbolic inputs represented as feature vectors, neural predicates typically operate on subsymbolic data and are not equipped to process structured symbolic inputs, such as graphs or relational structures, that would be required to fully capture the semantics of a knowledge base. This restricts \deepproblog{} to a fixed architecture of \texttt{subsymbol $\rightarrow$ neural $\rightarrow$ symbol}, limiting its flexibility in multi-layer reasoning scenarios involving structured data.

\deepgraphlog{} addresses both limitations by introducing \emph{graph neural facts}, which define dependencies between atoms using graph neural networks. Since atoms are probabilistic, the input to a GNN is itself a random graph (see \ref{sec:ProbLog}). This enables reasoning in the \texttt{symbol $\rightarrow$ neural $\rightarrow$ symbol} setting, even when explicit rules are missing. This key difference is illustrated in Figure \ref{fig:CircuitDiagram}, where \deepgraphlog{} can still perform reasoning without fully specified symbolic dependencies.

\textbf{Syntax} The \deepgraphlog{} language extends the syntax of \problog{} by introducing \emph{graph neural predicates}. A graph neural predicate serves as an interface between the logical and neural components: instead of assigning a fixed probability to a ground atom, the probability is computed by applying a Graph Neural Network (GNN) to a relational structure extracted from the possible world. 

\noindent
\smallskip
\textbf{Graph Neural Fact} 
A \textit{graph neural fact} is a ground instance of a graph neural predicate, and it has the following form:
\[
gnn(m_r, \gamma_r, [x_1,\dots,x_k])::r(x_1,\dots,x_k).
\]
where 
\begin{itemize}
    \item $gnn$ is a reserved function symbol -- we use $gnn$ for graph neural network;
    \item $m_r$ is a symbolic identifier that uniquely refers to a specific GNN architecture associated with the predicate $r$. Here, $m_r$ refers to a fixed architecture (e.g., the number and type of layers, the aggregation strategy, etc.) chosen for modelling $r$, while the actual parameters of the GNN (i.e., the weights) remain learnable;  
    \item $\gamma_r$ is a set of ground atoms that encodes the input graph. Unary atoms are interpreted as node attributes, and binary atoms as edges. If some of these atoms are probabilistic (as in \problog{}), then $\gamma_r$ does not represent a single graph but a distribution over graphs. In other words, $\gamma_r$ defines a \textit{random graph}, where each possible world $\omega_F$ yields a concrete instance $G = (\gamma_r \cap \omega_F)$ consisting of the atoms that are true in that world. The GNN then receives $G$ as its input. This aligns with the standard semantics of \problog{}, where each subset $F \subseteq \mathcal{F}$ induces a possible world, and the presence or absence of edges and node attributes in $G$ reflects the probabilistic nature of the facts in $\gamma_r$. Although $\gamma_r$ is formally defined as a ground set of atoms, in practice it can be specified using variables in a Prolog-style fashion, which are subsequently grounded during evaluation.
    \item $x_1,\dots,x_k$ are ground terms;
    \item $r(x_1,\dots,x_k)$ is the ground atom whose probability is computed by applying the $m_r$. 
\end{itemize}
Intuitively, a \textit{graph neural fact} extends a probabilistic fact whose probability is computed by a GNN model \(m_r\). 
However, notice that the set of atoms in $\gamma$ can be probabilistic facts, therefore, the GNN model has to handle input graphs that contain uncertainty.

\textbf{Example} We now introduce our first example of a \deepgraphlog{} program, based on the task shown in Figure \ref{fig:CircuitDiagram}. The objective of the task is to determine whether it is possible to \texttt{move} the block \texttt{a}. From the perspective of graph neural networks (GNNs), this is a node-level classification task, where the GNN \texttt{m\_move} learns to predict whether block \texttt{a} is movable based on its input graph (\texttt{on(a,b)} and \texttt{next\_to(a,c)}). To model this in the \deepgraphlog{} framework, we define the following program:

\begingroup
\footnotesize
\begin{minted}[frame=single]{prolog}
% input symbolic
0.7::on(a,b). 0.4::next_to(a,c). 0.5::light(a).
% reasoning neural
gnn(m_move,[on(a,b),next_to(a,c)],[a])::move(a).
\end{minted}
\endgroup

While we illustrate node-level classification task here, the notion of a \textit{graph neural predicate} serves as a general-purpose primitive for modelling a wide range of tasks on graphs, including edge and graph classification. For instance, a graph neural fact for edge classification can be expressed \texttt{gnn(m\textsubscript{livesIn}, $\gamma_{person}$, [x,y])::person(x,y).}, where \texttt{person(X,Y)} is an edge whose class is predicted. Similarly, binary graph classification can be modelled such as \texttt{gnn(m\textsubscript{classification}, $\gamma_{classification}$)::positive.}.

\textbf{Semantics} A \deepgraphlog{} program is defined as the tuple $(\mathcal{F}, \mathcal{R}, \mathcal{G})$, where $\cF$ is a set of ground probabilistic facts, $\cR$ is a set of rules, and $\cG$ is a set of ground graph neural facts.
The semantics of \deepgraphlog{} are defined in terms of ground programs. Since \deepgraphlog{} extends \problog{}, we describe its semantics in relation to the semantics of \problog{}.  
Similarly to \problog{}, a possible world $\omega$ is defined as the minimal model of a set of ground probabilistic facts, including both standard probabilistic facts $\mathcal{F}$ and graph neural facts $\mathcal{G}$, following the semantics described in Section \ref{sec:preliminaries}.  
Formally, each subset $F \subseteq (\mathcal{F} \cup \mathcal{G})$ defines a possible world $\omega_F$ given by $ \omega_{F} = F \cup \{ f\theta \mid \mathcal{R} \cup F \models f\theta \text{ and } f\theta \text{ is ground} \}.$

A \deepgraphlog{} program defines a probability distribution over possible worlds, similarly to \problog{} (eq. \ref{eq:ProbPossibleWorld}), but with conditioning to account for dependencies among facts. Given a graph neural fact $f \in \mathcal{G}$, let $\gamma_f$ denote the corresponding set of ground atoms that define its input random graph (see \ref{sec:ProbLog}). Since some atoms in $\gamma_f$ may be probabilistic, the actual input graph depends on which atoms are true in a given possible world. For a particular possible world $\omega_F$, we define the concrete input graph $G_f = \gamma_f \cap \omega_F$ as the subset of $\gamma_f$ that holds in $\omega_F$. Intuitively, each possible world induces a particular instantiation of the random graph $\gamma_f$, and this instance $G_f$ is the input graph presented to the neural predicate in that world.
The distribution is given by:
\begin{equation}\label{eq:ProbPossbileDGL}
    P(\omega_F) = \prod_{f \in F} P(f \mid G_f) \prod_{f \in (\mathcal{F} \cup \mathcal{G}) \setminus F} \big(1 - P(f  \mid G_f)\big),
\end{equation}
where
\[
P(\texttt{f} \mid G_f) =
\begin{cases}
m_r(G_f) & \text{if } f \in \mathcal{G}, \\
\texttt{p} & \text{if } \texttt{f} = \texttt{p::f},\ f \in \mathcal{F}.
\end{cases}
\]

\paragraph{\deepgraphlog{} Inference}
A query $q$ in \deepgraphlog{} is a ground atom for which we aim to compute the marginal probability $P(q)$ under the probabilistic semantics of the program.
\deepgraphlog{} supports two modes of inference depending on the structure of the query. If the query depends only on probabilistic facts $\cF$ and rules $\cR$ and not on any graph neural facts $\cG$ then inference reduces to standard \problog{} inference \cite{fierens2015inference}. Specifically, the marginal probability $P(q)$ can be computed using knowledge compilation \cite{darwiche2002knowledge} and it follows this setup: First, the set of all proofs $B(q)$ for the query is extracted via SLD resolution, compiled into a tractable circuit (e.g., SDD \cite{darwiche2011sdd}), and evaluated bottom-up using a semiring interpretation \cite{kimmig2017algebraic}. However, if the query includes graph neural facts $\cG$, the evaluation depends on the input graph $\gamma$, which varies per world and knowledge compilation is no longer applicable. Inference is then performed via explicit marginalization over all possible worlds: $P(q) = \sum_{\omega \models q , (\cF \cup \cG)} P(\omega),$
where the GNNs are evaluated for each world $\omega$ by constructing input graphs from $\omega$ as described in the semantics. This results in naive enumeration, which may be computationally expensive. However, in many practical applications, we expect the number of uncertain edges—and therefore the size of the search space—to remain limited. Future work will explore more efficient inference strategies, including approximate and sampling-based methods, as well as possible integration of knowledge compilation techniques.
    
\section{Experiments}
The goal of our experiments is to answer the following questions:
\begin{itemize}
    \item \textbf{Q1} Can \deepgraphlog{} combine learning and probabilistic reasoning to overcome the expressivity limitations of GNNs? 
    \item \textbf{Q2} How does \deepgraphlog{} compare to a graph neural network baseline w.r.t. accuracy and data efficiency?
    \item \textbf{Q3} Can \deepgraphlog{} be used for structure learning via parameter learning?
    \item \textbf{Q4} Can \deepgraphlog{} combine learning and reasoning to learn from distantly supervised examples?
    \item \textbf{Q5} Can graph neural predicates be applied for planning tasks?
\end{itemize}

\noindent
We used several experiments to evaluate \deepgraphlog. These experiments serve as an initial evaluation, showcasing the proof-of-concept of \deepgraphlog{}'s capabilities.

\begin{figure*}[t]
    \centering
    \begin{minipage}{0.45\textwidth}
        \centering
        \scalebox{0.7}{
            \begin{tikzpicture}[mynode/.style={circle, draw, inner sep=0.3pt, minimum size=0.35cm, fill=white, text=black, }]
    \node[mynode, label={[yshift=0] $[1]$}] (11) at (0,0) {\small a};
    \node[mynode, label={[yshift=0] $[1]$}] (12) at (1,0) {\small b};
    \node[mynode, label={[yshift=0] $[1]$}] (13) at (2,0) {\small c};   
    \node[mynode, label={[yshift=-2.7em] $[1]$}] (14) at (0,-1) {\small d};   
    \node[mynode, label={[yshift=-2.7em] $[1]$}] (15) at (1,-1) {\small e};   
    \node[mynode, label={[yshift=-2.7em] $[1]$}] (16) at (2,-1) {\small f};   

    \node[mynode, label={[yshift=0] $[1]$}] (21) at (3,0) {\small a};
    \node[mynode, label={[yshift=0] $[1]$}] (22) at (4,0) {\small b};
    \node[mynode, label={[yshift=0] $[1]$}] (23) at (5,0) {\small c};   
    \node[mynode, label={[yshift=-2.7em] $[1]$}] (24) at (3,-1) {\small d};   
    \node[mynode, label={[yshift=-2.7em] $[1]$}] (25) at (4,-1) {\small e};   
    \node[mynode, label={[yshift=-2.7em] $[1]$}] (26) at (5,-1) {\small f};   

    \node[mynode, label={[yshift=0] $[1]$}] (31) at (6,0) {\small a};
    \node[mynode, label={[yshift=0] $[1]$}] (32) at (7,0.5) {\small b};
    \node[mynode, label={[yshift=0] $[1]$}] (33) at (8,0) {\small c};   
    \node[mynode, label={[yshift=-2.7em] $[1]$}] (34) at (6,-1) {\small d};   
    \node[mynode, label={[yshift=-2.7em] $[1]$}] (35) at (7,-1.5) {\small e};   
    \node[mynode, label={[yshift=-2.7em] $[1]$}] (36) at (8,-1) {\small f};   
   
    \node at (1, 1.2) {$Y_{G_0} = 0$};
    \node at (4, 1.2) {$Y_{G_1} = 1$};
    \node at (7, 1.7) {$Y_{G_2} = 2$};
    
    \draw (11) -- (12);
    \draw (11) -- (14);
    \draw (12) -- (13);
    \draw (12) -- (15);
    \draw (13) -- (16);
    \draw (14) -- (15);
    \draw (15) -- (16);

    \draw (21) -- (22);
    \draw (21) -- (24);
    \draw (22) -- (24);
    \draw (22) -- (25);
    \draw (23) -- (25);
    \draw (23) -- (26);
    \draw (25) -- (26);

    \draw (31) -- (32);
    \draw (32) -- (33);
    \draw (33) -- (36);
    \draw (36) -- (35);
    \draw (35) -- (34);
    \draw (31) -- (34);

    \draw[->, bend right=10] (4, 2) to (1, 1.4); 
    \draw[->, bend left=10] (4, 2) to (4, 1.4);     
    
    \draw[->] (4, -2.5) to (4, -2);     
    \draw[->, bend right=10] (4, -2.5) to (6, -2); 

    \node (text1) at (4, 2.5)[align=center, font=\large] {Non-distinguishable by a 1-WL GNN. \\Distinguishable by the logical rule \texttt{class(0)}:-\texttt{cycle\_4.}};
    \node (text2) at (4, -3)[align=center,font=\large] {Distinguishable by a 1-WL GNN\\Non-distinguishable by the logical rule \texttt{class(0)}:-\texttt{cycle\_4.}};        
\end{tikzpicture}
        }        
        
        {\textbf{(a)} Logic at the top.}
        \label{fig:1WLEncodingLogicAtTop}
    \end{minipage}
    \hfill
    \begin{minipage}{0.45\textwidth}
        \centering
        \scalebox{0.7}{
            \begin{tikzpicture}[mynode/.style={circle, draw, inner sep=0.3pt, minimum size=0.35cm, fill=white, text=black}]    
    \node[mynode, label={[yshift=0] $[1,0]$}] (11) at (0,0) {\small a};
    \node[mynode, label={[yshift=0] $[1,0]$}] (12) at (1,0) {\small b};
    \node[mynode, label={[yshift=0] $[1,0]$}] (13) at (2,0) {\small c};   
    \node[mynode, label={[yshift=-2.7em] $[1,0]$}] (14) at (0,-1) {\small d};   
    \node[mynode, label={[yshift=-2.7em] $[1,0]$}] (15) at (1,-1) {\small e};   
    \node[mynode, label={[yshift=-2.7em] $[1,0]$}] (16) at (2,-1) {\small f};   
    \node[mynode, label={[yshift=-2.7em] $[0,1]$}] (17) at (0.5,-0.4) {\small g};   

    \node[mynode, label={[yshift=0] $[1,0]$}] (21) at (3.5,0) {\small a};
    \node[mynode, label={[yshift=0] $[1,0]$}] (22) at (4.5,0) {\small b};
    \node[mynode, label={[yshift=0] $[1,0]$}] (23) at (5.5,0) {\small c};   
    \node[mynode, label={[yshift=-2.7em] $[1,0]$}] (24) at (3.5,-1) {\small d};   
    \node[mynode, label={[yshift=-2.7em] $[1,0]$}] (25) at (4.5,-1) {\small e};   
    \node[mynode, label={[yshift=-2.7em] $[1,0]$}] (26) at (5.5,-1) {\small f};       
    \node[mynode, label={[yshift=-2.7em] $[0,0]$}] (27) at (4,-0.4) {\small g};   

    \node[mynode, label={[yshift=0] $[1,0]$}] (31) at (7,0) {\small a};
    \node[mynode, label={[yshift=0] $[1,0]$}] (32) at (8,0.5) {\small b};
    \node[mynode, label={[yshift=0] $[1,0]$}] (33) at (9,0) {\small c};   
    \node[mynode, label={[yshift=-2.7em] $[1,0]$}] (34) at (7,-1) {\small d};   
    \node[mynode, label={[yshift=-2.7em] $[1,0]$}] (35) at (8,-1.5) {\small e};   
    \node[mynode, label={[yshift=-2.7em] $[1,0]$}] (36) at (9,-1) {\small f};      
    \node[mynode, label={[yshift=-1.4em, xshift=-2em] $[0,0]$}] (37) at (8,-0.5) {\small g};   
   
    \node at (1, 1.2) {$Y_{G_0} = 0$};
    \node at (4.5, 1.2) {$Y_{G_1} = 1$};
    \node at (8, 1.7) {$Y_{G_2} = 2$};
    
    \draw (11) -- (12);
    \draw (11) -- (14);
    \draw (12) -- (13);
    \draw (12) -- (15);
    \draw (13) -- (16);
    \draw (14) -- (15);
    \draw (15) -- (16);
    \draw (11) -- (17);
    \draw (13) -- (17);
    \draw[-, bend left=45] (14) to (17); 
    \draw[-, bend right=45] (15) to (17);
    \draw[-, bend right=30] (16) to (17);

    \draw (21) -- (22);
    \draw (21) -- (24);
    \draw[-, bend right=40] (22) to (24);
    \draw (22) -- (25);
    \draw (23) -- (25);
    \draw (23) -- (26);
    \draw (25) -- (26);
    \draw (21) -- (27);
    \draw (22) -- (27);
    \draw (23) -- (27);
    \draw[-, bend left=45] (24) to (27);
    \draw[-, bend right=45] (25) to (27);
    \draw[-, bend right=30] (26) to (27);

    \draw (31) -- (32);
    \draw (32) -- (33);
    \draw (33) -- (36);
    \draw (36) -- (35);
    \draw (35) -- (34);
    \draw (31) -- (34);    
    \draw (31) -- (37);    
    \draw (32) -- (37);    
    \draw (33) -- (37);    
    \draw (34) -- (37);    
    \draw (35) -- (37);    
    \draw (36) -- (37);  

    \draw[->, bend left=10] (4, -3) to (4, -2);     
    \draw[->, bend right=10] (4, -3) to (7, -2); 

    \node (text2) at (4, -4)[align=center,font=\large] {Node $g$ has structural information encoded as $[0,0]$ \\ because no cycle of length 4 is present \\ in $G_1$ and $G_2$.};        
\end{tikzpicture}
        }        
        
        {\textbf{(b)} Logic at the bottom.}
        \label{fig:1WLEncodingLogicAtBottom}
    \end{minipage}

    \caption{Two different ways of encoding graph pattern of \texttt{treewidth > 1}. $Y_G$ represent the label of the graph and $[X]$ or $[X,Y]$ represent the node-level features. The task is to classify graphs based on the presence of a 4-cycle. Standard message-passing GNNs cannot detect due to limited expressivity. In logic-at-the-top, a GNN processes the graph, and the logical rule (e.g., \texttt{class(0) :- cycle\_4.}) is applied to its output. In logic-at-the-bottom, structural features like 4-cycles are encoded as new feature nodes connected to all original nodes, each with a unique binary vector. This allows the GNN to access symbolic structure through message passing and distinguish such patterns.}
    \label{fig:1WLEncoding}
\end{figure*}
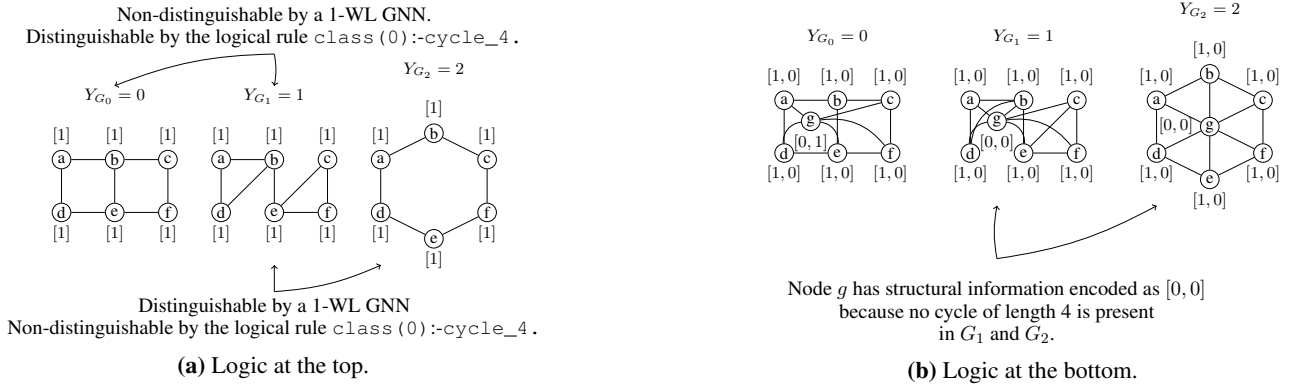

\textbf{E1: Beyond Weisfeiler-Leman}
Standard message-passing GNNs cannot distinguish certain graph structures—for example, $k$-regular graphs of the same size—due to an expressivity limited to that of 1-WL isomorphism test \cite{Xu+2018}. Many architectures have been proposed to enhance GNN expressivity. In this experiment, we focus specifically on the expressivity of 1-WL GNNs. The goal is to show that logical knowledge can help distinguish graphs that GNNs alone cannot, while GNNs can still differentiate graphs when explicit logical rules are unavailable. We compare two approaches:

\textit{Logic-at-the-top}: This approach integrates logical reasoning as a separate layer above the GNN. Instead of injecting symbolic knowledge into the node features directly, logical rules are used to reason about the outputs of the GNN. This setting is naturally expressed in \deepgraphlog{}, where one can combine GNNs for learning over relational graph structures with symbolic rules. An illustration of this approach is shown in Figure \ref{fig:1WLEncoding} (a).

\textit{Logic-at-the-bottom}: Inspired by \cite{bouritsas2022improving,kikaj2024subgraph}, this baseline identifies graph properties that 1-WL GNNs cannot capture and encodes each of such structure as a new node that is added to the graph. Each feature node is connected to all existing nodes and is annotated with a unique feature type via its feature vector. This augmentation encodes symbolic information to the GNN through message passing, allowing us to test whether GNNs can leverage such explicitly provided relational cues. An illustration of this encoding is shown in Figure \ref{fig:1WLEncoding}(b).

\begin{figure*}[t]
    \centering
    \begin{minipage}{0.3\textwidth}
        \centering
        \includegraphics[width=\linewidth]{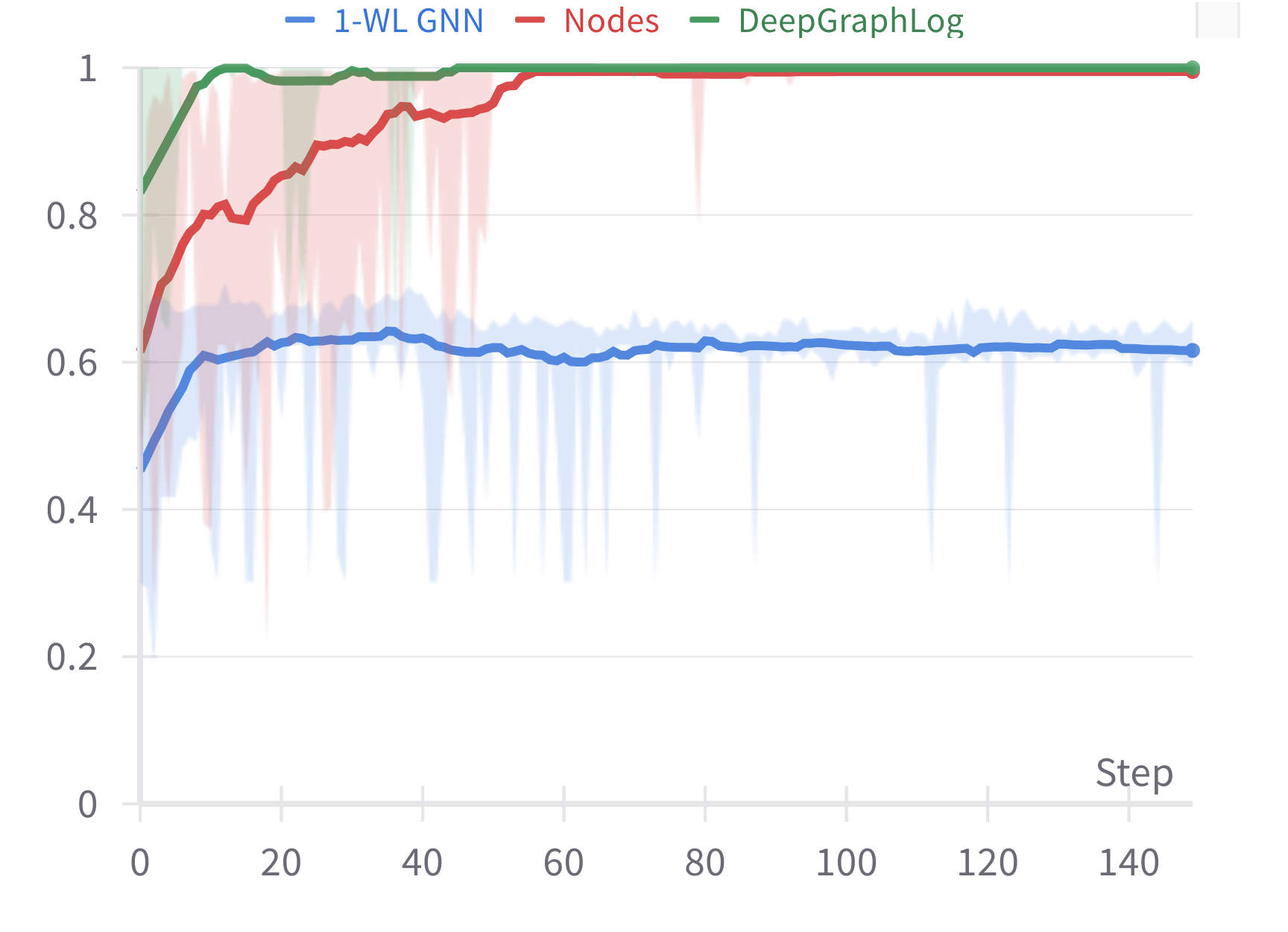}
        {\textbf{(a)} Test set accuracy — Trained on 100\% of training set}
        \label{fig:T1A}
    \end{minipage}
    \hfill
    \begin{minipage}{0.3\textwidth}
        \centering
        \includegraphics[width=\linewidth]{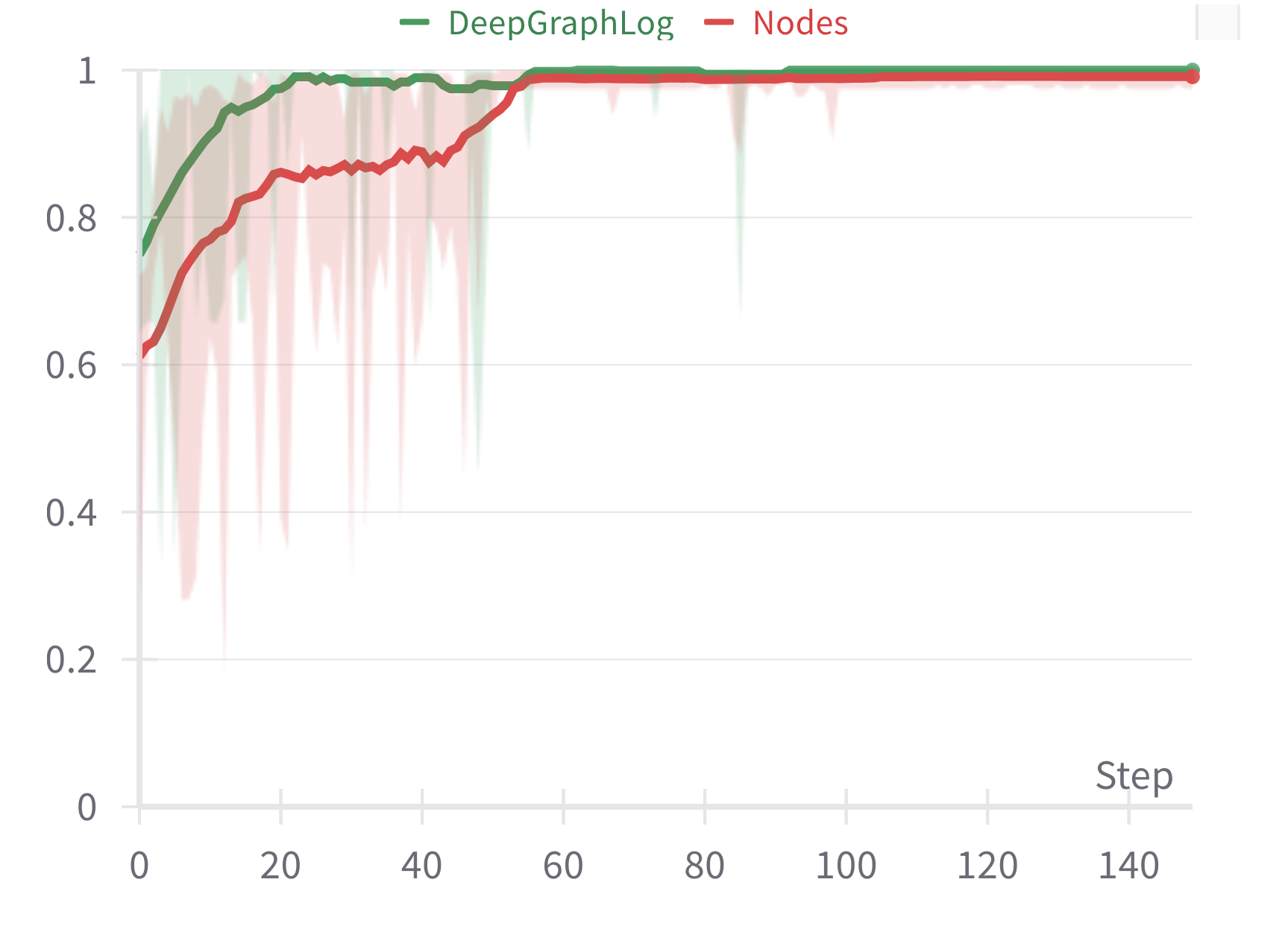}
        {\textbf{(b)} Test set accuracy — Trained on 50\% of training set}
        \label{fig:T2A}
    \end{minipage}
    \hfill
    \begin{minipage}{0.3\textwidth}
        \centering
        \includegraphics[width=\linewidth]{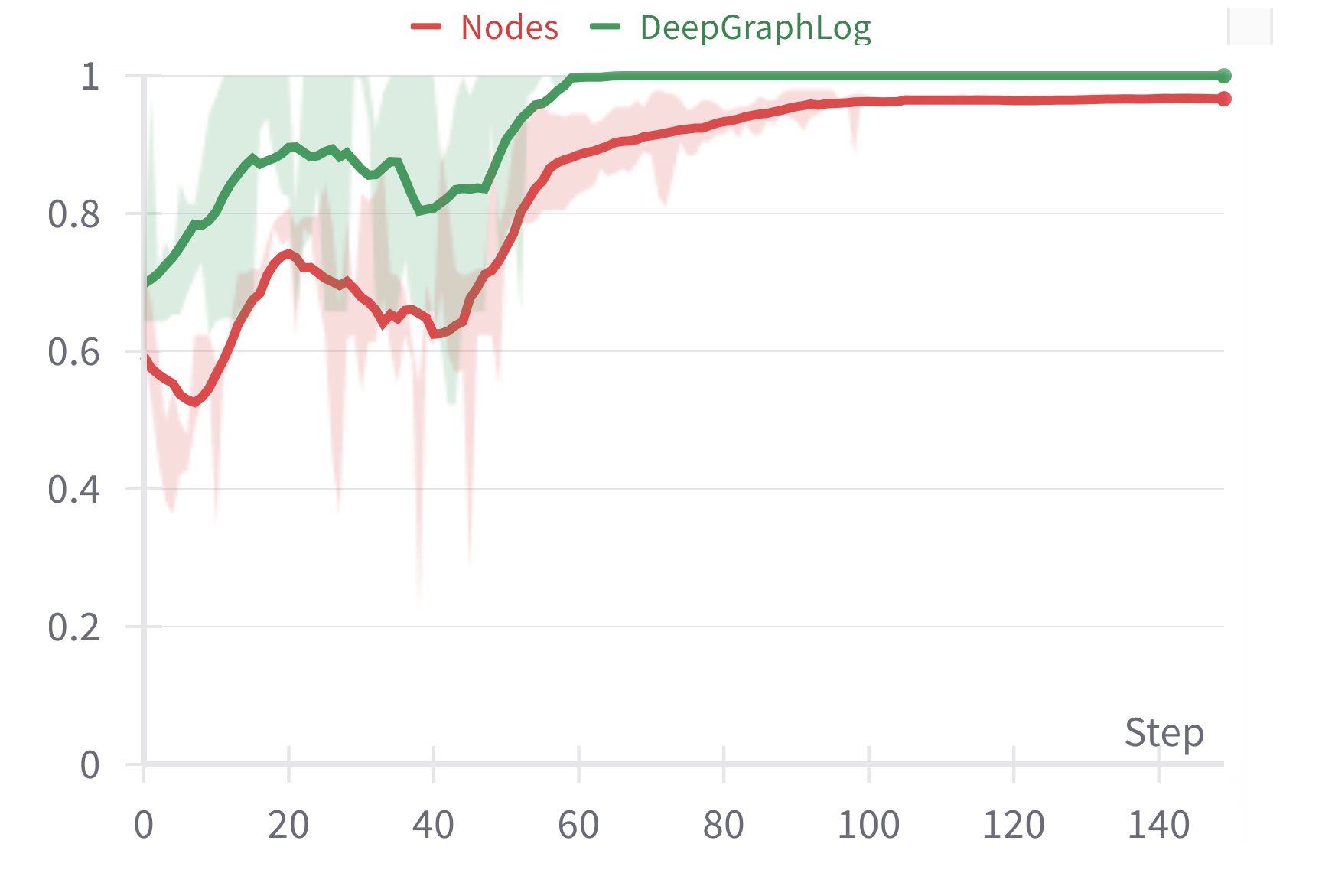}
        {\textbf{(c)} Test set accuracy — Trained on 10\% of training set}
    \end{minipage}

    \caption{\deepgraphlog{} for GNN expressivity (T1). The graphs show the accuracy on the test set during training. \deepgraphlog{} converges quicker and to a higher accuracy than the baseline, especially with fewer examples.}
    \label{fig:T1Results}
\end{figure*}

\textbf{E2: Structure Learning }
Building upon the previous task, this experiment evaluates the ability of \deepgraphlog{} to perform structure learning via parameter learning \cite{marra2024statistical}. Instead of explicitly encoding the logical rules that state how the discriminative patterns contribute to the classification, the model must learn the importance of different graph structures for a given downstream task. In this setting, we assume that a set of candidate graph structures (templates) that could be relevant for reasoning are given. Specifically, we provide the model with a set of candidate graph structures (\texttt{c1,c2,...,cn}) such as cycles or cliques \cite{bouritsas2022improving}. However, since we do not know which structures contribute to the prediction, we assign each structure a probability (as a probabilistic fact) that is learned jointly with the task (i.e., graph classification). 
For example, to classify a graph $G$ as belonging to \texttt{class\_0}, our model follows these rules (\texttt{rF(X,Y)} is a shorthand notation for \texttt{relevantFor(X,Y)}):

\begingroup
\footnotesize
\begin{minted}[frame=single]{prolog}
p1::rF(c1, class_0). p2::rF(c2, class_0). ...
classify(G,class_0) :- gnn_classifier(G, class_0).
classify(G,class_0) :- rF(c1, class_0).
classify(G,class_0) :- rF(c2, class_0). ...
\end{minted}
\endgroup
In this formulation, a graph $G$ is classified as \texttt{class\_0} if the GNN predicts it as such or if it contains a particular structural template, that has a probability assigned to it (e.g., \texttt{p1::rF(c\_1, class\_0).}). By learning these probabilities, the model can automatically determine which structural features are most relevant to the classification task.

\textbf{E3: Distant Supervision for Knowledge-Graph Completion}
To demonstrate that \deepgraphlog{} supports learning in knowledge graphs with only partial logic knowledge and distant supervision, we use the family-tree graph from \cite{hohenecker2020ontology}, which is similar to the well-known Kinship dataset \cite{kinship_55}. In this task, the knowledge graph consists of nodes labelled as \texttt{person(X)} and their gender, either \texttt{male(X)} or \texttt{female(X)}. The only given input relationship (graph) is \texttt{parentOf(X,Y)}, and our goal is to learn two additional relationships: \texttt{fatherOf} and \texttt{motherOf} (i.e., we want to predict the links \texttt{motherOf} and \texttt{fatherOf}). However, we are not provided with any label on such links. On the contrary, we are provided with a distant supervision signal on whether people are the grandfather of each other along with the following background knowledge.

\begingroup
\footnotesize
\begin{minted}[frame=single]{prolog}
% graph = m -> male, f -> female, pOf -> parentOf
gnn(gcn_fOf,[m/1,f/1,pOf/2],[X,Y])::fatherOf(X,Y).
gnn(gcn_mOf,[m/1,f/1,pOf/2],[X,Y])::motherOf(X,Y).
grandfatherOf(X,Y) :- fatherOf(X,Z),fatherOf(Z,Y).
grandfatherOf(X,Y) :- fatherOf(X,Z),motherOf(Z,Y).
\end{minted}
\endgroup

Notice that a pure logic system cannot use the rules as it has no information about what being a mother or being a father mean. On the contrary a graph neural network alone can learn the grandfather relation but has no way to trace it back to being a father or mother. \deepgraphlog{} bridges the two. To demonstrate that this task can be effectively solved with \deepgraphlog{}, we model the relationships \texttt{fatherOf} and \texttt{motherOf} using a \textit{graph neural predicate} for each of the relationships. As a baseline, we also train a standard GNN model to directly predict the \texttt{fatherOf} and \texttt{motherOf} relations using the same input graph. However, this model receives no direct supervision for these predicates, and the only available training signal comes from the \texttt{grandfatherOf} predicate.

\textbf{E4: Blocks World — Multi-Layer Reasoning} To showcase the key capability of \deepgraphlog{} — namely, multi-layer reasoning — we propose a planning task inspired by the classic \textit{Blocks World} domain \cite{slaney2001blocks}. The task involves deciding whether a valid tower can be formed in a single move, given an initial configuration of blocks on a floor.

Blocks can be of type \texttt{metal}, \texttt{plastic}, or \texttt{glass}, and a key constraint encoded in the symbolic logic is that no block can be moved onto a glass block: 

\begingroup
\footnotesize
\begin{minted}[frame=single]{prolog}
illegal(X,Y):-move(X,Y),glass(Y).
\end{minted}
\endgroup

\begin{figure*}[t]
    \centering
    \includegraphics[width=0.95\linewidth]{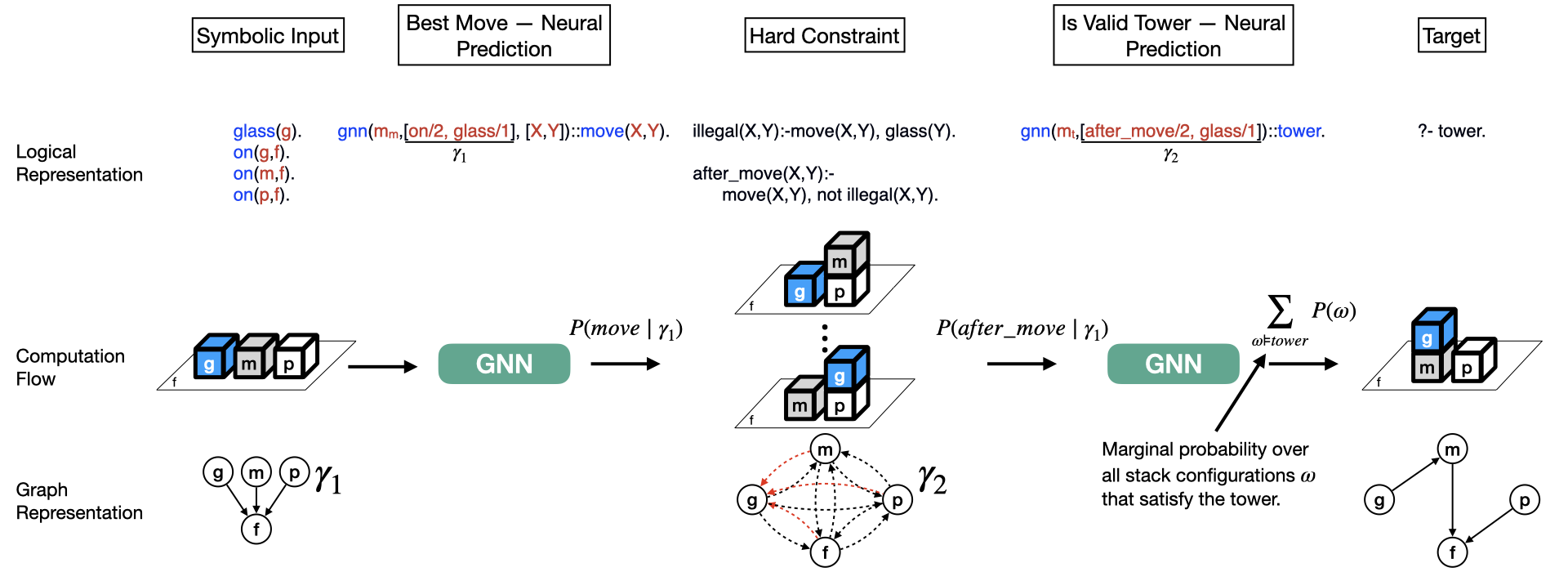}        
    \caption{Blocks World Inspired Task Represented as a Multi-Layer Reasoning Task in \deepgraphlog.}\label{fig:MultiLayerTask}
\end{figure*}

The input is a symbolic state describing the blocks and their initial positions (see Fig. \ref{fig:MultiLayerTask}). This state is processed by a graph neural predicate \texttt{move(X,Y)} that predicts possible moves between blocks. The predicted moves are filtered using the hard logic constraint \texttt{illegal(X,Y)} to remove invalid ones. The resulting legal moves define a transition state, encoded as \texttt{after\_move(X,Y)}, which is then passed to a second graph neural predicate \texttt{tower}.

The second GNN receives this updated relational structure and predicts whether the resulting configuration forms a valid tower. In this task, a valid tower is any stacking configuration where at least two blocks are stacked and the topmost block is made of glass.

We compare against two alternative baselines: (i) a single GNN that directly predicts whether a valid tower can be formed from a given configuration, and (ii) a two-GNN model without applying the hard constraint between layers (denoted “GNN w/o Constraints”).

\textbf{Results} For experiments \textbf{E1, E2} and \textbf{E4}, we report the standard deviation over 5 runs.

\textbf{E1:} To answer \textit{Q1} and \textit{Q2}, we compare \deepgraphlog{} (logic-at-the-top) with a standard 1-WL GNN and a GNN (baseline) where the graph data receive discriminative node features (logic-at-the-bottom). We evaluate all three models on a graph classification task under varying data availability: 1000, 500, and 100 examples (Figure \ref{fig:T1Results}). For $N = 1000$ and $N = 500$ (Figures \ref{fig:T1A} (a) and \ref{fig:T2A} (b)), \deepgraphlog{} converges faster than the baseline but with comparable accuracy. At $N = 100$, however, it outperforms the baselines in both convergence and final accuracy. While the baseline must learn how to use the input features,  \deepgraphlog{} can directly encode and reason over them. When the role of such features is unclear, they can still be included as input—as explored in the next experiment.

\textbf{E2:} To answer \textit{Q3}, our results shown in Table \ref{tab:structure_learning} that \deepgraphlog{} can effectively perform structure learning through parameter learning. Specifically, the model assigns a probability of $1.0$ to the template \texttt{cycle\_4} being relevant for \texttt{class\_0}, while assigning $0.0$ to all other classes—matching the true discriminative pattern. This indicates that \deepgraphlog{} successfully identifies and exploits the most relevant structural feature without it being hardcoded. Furthermore, it achieves higher classification accuracy than standard GNNs, and closely approximates the performance of a version of the program where the relevant structures and their probabilities are explicitly provided. These results demonstrate that learning structural patterns, rather than relying solely on node-level features, yields a significant advantage in tasks requiring symbolic abstraction.

\begin{table}[t]
\centering
\caption{T2: learned probabilities for \texttt{cycle\_4} and \texttt{cycle\_3} across classes.}
\label{tab:structure_learning}
\begin{tabular}{lccc}
\toprule
\textbf{Class} & \texttt{class\_0} & \texttt{class\_1} & \texttt{class\_2} \\
\midrule
\texttt{P(rF(c4, c0))} & $1.0$ & $0.0$ & $0.0$ \\
\texttt{P(rF(c3, c1))} & $0.0$ & $0.8$ & $0.2$ \\
\bottomrule
\end{tabular}
\end{table}

\textbf{E3:} To answer \textit{Q4}, the results in Table \ref{table:resE2} show that \deepgraphlog{} (DGL) significantly outperforms the standard GNN baseline on both \texttt{fatherOf} and \texttt{motherOf}. For \texttt{fatherOf}, DGL achieves an F1 score of 98.94 compared to 22.95 for the GNN, and for \texttt{motherOf}, 98.96 versus 59.61. \deepgraphlog{} also reaches perfect or near-perfect scores across all other metrics— 
AUCROC, and Hits@K—while the GNN remains below $65\%$ in every case. These results confirm that \deepgraphlog{} effectively leverages distant supervision and background knowledge to learn intermediate relations that a GNN alone cannot capture.

\begin{table}[t]
\label{table:resE2}
\caption{Performance on T3: learning intermediate relations \texttt{fatherOf} and \texttt{motherOf} from distant supervision.}
\centering
\begin{tabular}{lcccc} 
\toprule
\textbf{Relation} & \textbf{Metric} & \textbf{GNN} & \textbf{DGL} \\
\midrule
\multirow{5}{*}{\texttt{fatherOf}} 
    & F1        & $40.80\pm29.04$ & $98.28\pm1.47 $ \\
    & AUCROC    & $53.40\pm3.78$ & $99.23\pm0.09 $ \\
    & Hits@5    & $52.00\pm10.95  $ & $100.00\pm0.00$ \\
    & Hits@20   & $53.00\pm2.74  $ & $100.00\pm0.00$ \\
\midrule
\multirow{5}{*}{\texttt{motherOf}} 
    & F1        & $59.60\pm9.69 $ & $98.37\pm1.32$ \\
    & AUCROC    & $48.60\pm4.56 $ & $99.22\pm0.10$ \\
    & Hits@5    & $56.00\pm21.91$ & $100.00\pm0.00$ \\
    & Hits@20   & $51.00\pm5.48$ & $100.00\pm0.00$ \\
\bottomrule
\end{tabular}
\end{table}

\textbf{E4:} To answer \textit{Q5} — the results of Experiment 4 (Table \ref{table:resE4}) show that \deepgraphlog{} provides a clear advantage in such settings. The baseline GNN, which directly predicts whether a valid tower can be formed, achieves only $65.99\%$ accuracy, indicating difficulty in learning the compositional structure of the task. Introducing a second GNN without constraints (GNN w/o Constraints) improves performance to $75.44\%$, suggesting that modular reasoning helps. However, \deepgraphlog{}, which combines two graph neural predicates with an explicit symbolic constraint in between, achieves perfect accuracy ($100.0\%$). This demonstrates that graph neural predicates, when combined with logical structure, can successfully handle planning tasks that require relational reasoning and symbolic enforcement. These results highlight the importance of structured intermediate representations and hard constraints in enabling effective and interpretable planning with neural components.

\begin{table}[t]
\label{table:resE4}
\caption{
T4: Comparison of Different Architectures on the  Planning Task.}
\centering
\begin{tabular}{lccc} 
\toprule
Task & \textbf{GNN} & \textbf{GNN w/o Constraints} & \textbf{\deepgraphlog{}} \\
\midrule
T4 & $65.99 \pm 1.79$ & $75.44 \pm 3.00$ & $100.0 \pm 0.0$ \\
\bottomrule
\end{tabular}
\end{table}

\section{Related Work}

Most work on combining neural networks and logical reasoning in neurosymbolic AI targets knowledge graphs \cite{delong2024neurosymbolic}. These models typically follow a one-way \texttt{neural $\rightarrow$ symbol} structure, where logic is applied only after neural processing. An early example is BetaE \cite{Ren+2020}, which learns Beta distributions over embeddings to support multi-hop and negated queries via probabilistic set representations. GNN-QE \cite{zhu2022neural} instead uses GNNs for relation projection and fuzzy logic for query composition, treating each logical operation separately. BetaE offers efficiency by embedding full queries into a single distribution, while GNN-QE prioritizes interpretability through modular reasoning. Other approaches, such as ExpressGNN \cite{zhang2020efficient}, use GNNs as inference engines within probabilistic frameworks like Markov Logic Networks. LRNNs \cite{vsourek2021beyond} and SATNet \cite{wang2019satnet} can in principle allow for layered learning and reasoning, but are limited by their expressivity (e.g. Datalog)
and their semantics (i.e. fuzzy logic relaxation of hard constraints \cite{van2022analyzing}). In contrast, \deepgraphlog{} supports two-way interaction between neural and symbolic components, allowing for \texttt{neural $\rightarrow$ symbol $\rightarrow$ neural} setting. Built on probabilistic logic programming, it combines soft and hard constraints under well-defined semantics, improving correctness and interpretability. 
\deepgraphlog{} remains fully differentiable and handles complex graph data, making it well-suited for reasoning tasks requiring both structure and flexibility.

\section{Conclusion}
We introduced \deepgraphlog{}, a new neurosymbolic framework that integrates Graph Neural Networks (GNNs) with probabilistic logic programming enabling multi-layer neural-symbolic reasoning. Unlike existing approaches 
enforcing a fixed layering of neural and symbolic components, \deepgraphlog{} allows arbitrary interleaving of neural and symbolic layers, thus enhancing expressivity and reasoning capabilities. Through experiments on GNN expressivity, knowledge graph completion, and planning domain, we demonstrated the flexibility of \deepgraphlog{} and its power to learn relational structures, generalizing beyond the Weisfeiler-Leman (WL) expressivity limit, and incorporating background knowledge to guide learning.
Despite the advantages, \deepgraphlog{} introduces new challenges, including increased computational complexity due to recursive dependencies between neural and symbolic components.
Future work will focus on optimizing inference efficiency, and expanding the framework to more complex reasoning tasks such as multi-step planning.

\begin{ack}
This research received funding from the KU Leuven Research Fund (iBOF/21/075), from the Flemish Government under the ”Onderzoeksprogramma Artificiele Intelligentie (AI) Vlaanderen” programme, from the Wallenberg AI, Autonomous Systems and Software Program (WASP)
funded by the Knut and Alice Wallenberg Foundation, and from the European Research Council (ERC) under the European Union’s Horizon Europe research and innovation programme (grant agreement n$^\circ$101142702).
\end{ack}

\bibliography{mybibfile}

\end{document}